\documentclass[10pt,journal,compsoc]{IEEEtran}
\usepackage{algorithm}
\usepackage{algorithmic}
\usepackage{multicol}
\usepackage{multirow}
\usepackage{booktabs}
\usepackage{graphicx}
\usepackage{amsmath}
\usepackage{amsfonts}
\usepackage{colortbl}  
\usepackage{xcolor}

\ifCLASSOPTIONcompsoc
  \usepackage[nocompress]{cite}
\else
  \usepackage{cite}
\fi
\ifCLASSINFOpdf
\else
\fi
\begin{document}
%
\title{One Subgraph for All: Efficient Reasoning on Opening Subgraphs for Inductive Knowledge Graph Completion}
%
%
%
%

\author{Zhiwen Xie,
        Yi Zhang, Guangyou Zhou, Jin Liu, Xinhui Tu,  
        and~Jimmy Xiangji Huang,~\IEEEmembership{Member,~IEEE}
\IEEEcompsocitemizethanks{\IEEEcompsocthanksitem Zhiwen Xie,Guangyou Zhou and Xinhui Tu are with the School of Computer Science, Central China Normal University, Wuhan, 430079, China. E-mail: \{zwxie@ccnu.edu.cn; gyzhou@mail.ccnu.edu.cn; tuxinhui@mail.ccnu.edu.cn\}
\IEEEcompsocthanksitem Yi Zhang is with the Faculty of Artificial Intelligence in Education, Central China Normal University, Wuhan, 430079, China. E-mail: yizhang@mails.ccnu.edu.cn
\IEEEcompsocthanksitem Jin Liu is with the School of Computer Science, Wuhan University, Wuhan, 430072, China. E-mail: jinliu@whu.edu.cn
\IEEEcompsocthanksitem Jimmy Xiangji Huang is with the Information Retrieval and Knowledge Management Research Lab, York University, Toronto, Canada. E-mail: jhuang@yorku.ca.}
\thanks{Manuscript received April 19, 2005; revised August 26, 2015.}}

%
%

\markboth{Journal of \LaTeX\ Class Files,~Vol.~14, No.~8, August~2015}%
{Shell \MakeLowercase{\textit{et al.}}: Bare Demo of IEEEtran.cls for Computer Society Journals}
%



\IEEEtitleabstractindextext{%
\begin{abstract}
Knowledge Graph Completion (KGC) has garnered massive research interest recently, and most existing methods are designed following a transductive setting where all entities are observed during training. Despite the great progress on the transductive KGC, these methods struggle to conduct reasoning on emerging KGs involving unseen entities. Thus,  inductive KGC, which aims to deduce missing links among unseen entities, has become a new trend. Many existing studies transform inductive KGC as a graph classification problem by extracting enclosing subgraphs surrounding each candidate triple. Unfortunately, they still face certain challenges, such as the expensive time consumption caused by the repeat extraction of enclosing subgraphs, and the deficiency of entity-independent feature learning. To address these issues, we propose a global-local anchor representation (GLAR) learning method for inductive KGC. Unlike previous methods that utilize enclosing subgraphs, we extract a shared opening subgraph for all candidates and perform reasoning on it, enabling the model to perform reasoning more efficiently. Moreover, we design some transferable global and local anchors to learn rich entity-independent features for emerging entities. Finally, a global-local graph reasoning model is applied on the opening subgraph to rank all candidates. Extensive experiments show that our GLAR outperforms most existing state-of-the-art methods.
\end{abstract}

\begin{IEEEkeywords}
Knowledge Graph, Inductive Reasoning, Link Prediction, Anchor, Subgraphs.
\end{IEEEkeywords}}

\maketitle

\IEEEdisplaynontitleabstractindextext

%
\IEEEpeerreviewmaketitle

\section{Introduction}\label{}
\IEEEPARstart{K}{nowledge} graphs (KGs) organize real-world knowledge in a structural form of factual triple $(h,r,t)$, which means that the head entity $h$ is connected with the tail entity $t$ via the relation $r$. 
Recent years have witnessed great progress on large-scale KGs such as YAGO \cite{SuchanekKW2007}, Freebase \cite{Freebase}, DBPedia \cite{DBPedia} and Wikidata \cite{Wikidata}. These KGs have shown promising potential to support many downstream tasks, including but not limited to semantic search \cite{XiongPC17,ZhangJD0YCTHWHC21}, question answering \cite{YasunagaRBLL21,LiX22}, dialog generation \cite{WanSYQB23,RonyU022} and many more. Despite the large amounts of facts in KGs, it remains a challenge that KGs suffer from an incompleteness issue, since lots of links are missing. Therefore, knowledge graph completion (KGC), which is dedicated to infer missing links in KGs (thus also known as link prediction), has attracted increasing research attention.

\begin{figure*}[t]
\centering
\includegraphics[width=0.85\textwidth]{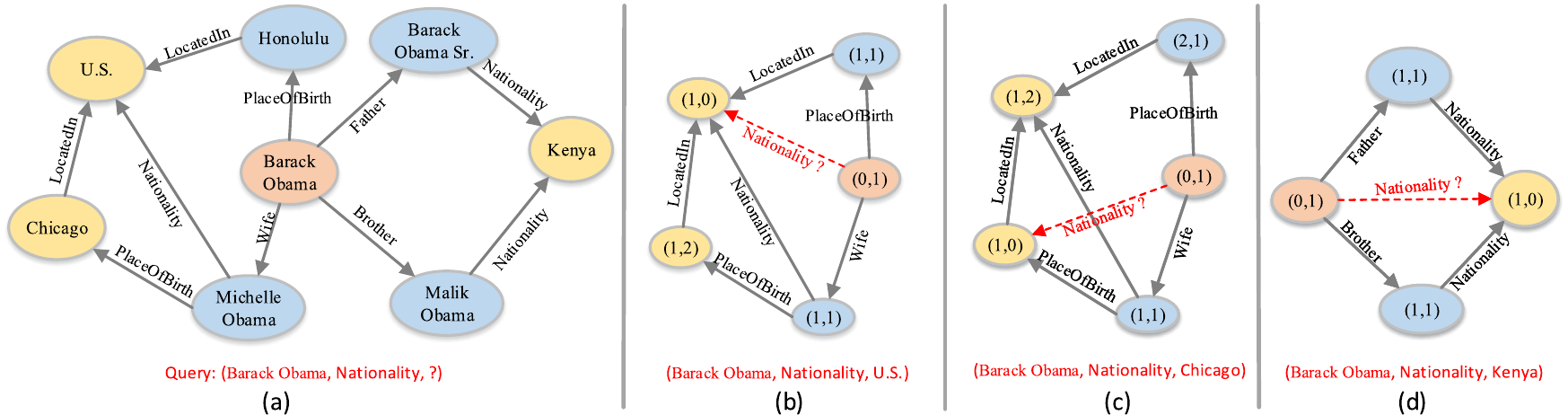}
\caption{An example of enclosing subgraphs extracted by previous inductive KGC methods, such as GraIL \cite{GraIL}, CoMPILE \cite{CoMPILE}, TACT \cite{TACT}, ConGLR \cite{ConGLR}. The query is ({\em Barack Obama}, {\em Nationality}, ?) and the subfigure (a) is a subgraph surrounding the query. The subfigures (b), (c) and (d) depict three enclosing subgraphs extracted for different candidates, where the nodes are labeled with the distances from each node to the entities in each candidate link.}
\label{fig_example}
\end{figure*}

Many typical KGC models are proposed to learn entity and relation embeddings and predict missing relations between entities, such as TransE \cite{TransE}, RotatE \cite{RotatE}, CompGCN \cite{CompGCN}, R-GCN \cite{RGCN}, KGTuner \cite{KGTuner}, ReInceptionE \cite{ReInceptionE}, PairRE \cite{PairRE} and so on.  Although these methods achieve satisfactory performance on KGC task, they all follow a {\bf transductive setting} in which all entities in KGs can be observed during training. However, KGs are continuously evolving with many new entities emerging daily in real world. Traditional KGC models with the transductive setting are not able to perform reasoning on new KGs with emerging entities unless retraining the whole KGs from scratch.
This prompts researchers to investigate models for KGC methods under an {\bf inductive setting}, where missing links can be inferred in a new KG composing of unseen entities. It is more challenging than conventional transductive KGC, since the entities in training and inference phrase are totally different.

Recently, numerous models for inductive KGC have been developed, including rule-based and GNN-based models. Rule-based approaches, such as RuleN \cite{RuleN}, DRUM \cite{DRUM} and NeuralLP \cite{NeuralLP}, mine logical rules by counting the co-occurrence patterns of relations on training KGs, and then apply these rules to infer missing links on test KGs. Since these logical rules are composed of entity-independent relations, they can be naturally generalized to testing KGs with new entities. Despite the inherent inductive reasoning capability, their performance is unsatisfactory due to paucity of meaningful rules. GNN-based methods include GraIL \cite{GraIL} and its extensions, such as TACT \cite{TACT},  CoMPILE \cite{CoMPILE}, ConGLR \cite{ConGLR} and Meta-iKG \cite{MetaiKG}. These methods transform the inductive link prediction task into a graph classification problem, which need to extract an enclosing subgraph around each target candidate triple and then apply graph neural networks (GNNs) on all candidate subgraphs. Although these GNN-based methods have shown promising improvement on inductive KGC, they suffer from the problems of inefficient reasoning and indistinguishable node labeling.

An example is illustrated in Figure~\ref{fig_example}, where the query is ({\em Barack Obama}, {\em Nationality}, ?). For different candidate triples, previous enclosing subgraph based methods will extract different enclosing subgraphs as shown in subfigures (b), (c) and (d). We can see that the subgraphs (b) and (c) have the same entity set and structure, while previous methods have to repeatedly extract two enclosing subgraphs and perform reasoning on these two subgraphs, which will greatly increase the consumption of reasoning time.  In addition, previous methods label each node in subgraphs with only distance features, failing to capture expressive node features. As illustrated in Figure~\ref{fig_example}, both the entity {\em Michelle Obama} in subfigure (c) and the entity {\em Malik Obama} in subfigure (d) are labeled as "(1,1)", leading to similar node features for nodes in different subgraphs. This simple node labeling strategy limits the model's ability to learn sufficient entity-independent node features for better inductive reasoning. Therefore, it becomes an urgent need to develop a new inductive reasoning paradigm to perform inductive link prediction efficiently and mine adequate transferable entity-independent anchors to effectively learn embeddings for emerging entities.  

Motivated by this, previous work QAAR \cite{QAAR} proposes a query adaptive anchor representation method \cite{QAAR} to address these issues of existing GNN-based methods. This preliminary study uses an approach to extract an opening subgraph that covers all candidate entities for a given query, {\bf enabling the model to rank all candidates using one subgraph}. Moreover, the QAAR model devises some query-aware entity-independent anchors to better capture the feature vector of nodes in subgraphs. However, this preliminary work still suffers from certain limitations in learning expressive features for nodes far away from the center node of an opening subgraph. Since the defined query-aware anchors are locally distributed around the query, resulting in insufficient anchors to learn structure features for distant nodes, especially for the nodes outside of opening subgraphs.  To address these challenges, we introduce a new global-local anchor representation (GLAR) learning method for inductive KGC in this study, which leverages both local and global transferable entity-independent anchors to learn inductive embeddings. The local transferable anchors are defined as the center node and its one-hop neighbors, and the global anchors are selected by using a clustering algorithm. Specifically, we use the neighboring relations of nodes as node features for clustering and perform clustering over these node features. The nodes which are closest to the clustering centroids are selected as global anchors. Then, we can label each node based on these global and local anchors to learn rich entity-independent features. Finally, we apply a global-local graph reasoning model to collaboratively propagate both local and global neighborhood features. Comprehensive experiments on three commonly used inductive KGC datasets are conducted and our GLAR model outperforms other state-of-the-art models on the inductive KGC task.

In summary, we briefly list the contributions as follows:
\begin{itemize}
    \item We develop a novel paradigm for inductive KGC by extracting opening subgraph and performing reasoning on it, which is more efficient than previous enclosing subgraph based methods.
    \item We design local and global anchors to learn rich entity-independent structure information to enhance the representations of entities in KGs.
    \item We propose a new global-local graph reasoning model to aggregate both local and global features in subgraphs.
    \item Extensive experiments demonstrate the effectiveness and efficiency of the proposed GLAR model.
\end{itemize}


\section{Related Work}
\label{sec_related_work}
\subsection{Transductive KGC }
Over the past decade, numerous techniques for transductive KGC have been introduced to address the incompleteness of KGs, including geometry-based methods, tensor decomposition-based, deep learning-based and graph-based approaches. Geometry-based methods treat the head and tail entities as two vectors in geometry space, and the head entity can be transformed to the tail entity via a relation-specific translation or rotation operation, such as TransE\cite{TransE}, TransH\cite{TransH}, TransR\cite{TransR}, RotatE\cite{RotatE}, PairRE\cite{PairRE} and HAKE\cite{HAKE}.   Decomposition-based methods are also known as bilinear models, such as Rescal \cite{RESCAL}, DistMult \cite{DistMult}, ComplEx \cite{ComplEx}. They model a KG as a 3D tensor where each element represents the credibility score of a corresponding triple. The relation and entity embeddings are learned by decomposing the 3D tensor of a KG. These geometry-based models and decomposition-based models employ shallow neural networks to learn KG embeddings, which can be limited in their expressive capacity. Therefore, some deep learning-based models are suggested to capture more expressive features, such as ConvE \cite{ConvE}, ConvR\cite{ConvR}, InteractE\cite{InteractE}, ReInceptionE\cite{ReInceptionE}. More recently, several graph-based models are introduced to capture neighborhood information in KGs, including R-GCN\cite{RGCN}, CompGCN \cite{CompGCN}, DisenKGAT\cite{DisenKGAT}. Unfortunately, these methods all follow the transductive setting where the full set of entities and relations require to be observed during training, while the emerging new entities and relations cannot be learned.

\subsection{Inductive KGC}
In real world, KGs constantly evolve with new entities continually added, such as new users and items which are crucial in recommendation systems \cite{LeeIJCC19,WuZGYZ21,Qian2019,Zhong2018}. Therefore, inductive KGC, which aims to infer missing facts in KGs with emerging entities and relations, has garnered increasing attention. Inductive KGC methods can be broadly classified into three categories \cite{DEKGILP}, namely text-based, rule-based and GNN-based methods.

\subsubsection{Text-based Methods}
In inductive KGC, it is a key problem to learn embeddings for emerging entities in new KGs. Some studies take advantage of additional textual descriptions to learn representations for entities. Thus, the representations for unseen entities can be generated based on their textual descriptions. KG-BERT\cite{KGBERT} proposed a BERT based model which feeds the textual descriptions of head, relation and tail into the BERT model and predicts the score for a triple using BERT\cite{BERT}  model. KEPLER \cite{KEPLER} presented a knowledge-enhanced language model by jointly learning knowledge embedding and language modeling objective. BERTRL \cite{BERTRL} incorporates the rule path into BERT model to improve the explainability. SimKGC \cite{SimKGC} uses contrastive learning to learn better entity embeddings. However, these methods rely on rich textual descriptions as additional information which is not always available in practical applications. In contrast, situations without using any additional textual information are more general and more challenging in various real-world applications \cite{MorsE}. Therefore, we do not use additional textual information in this study.

\subsubsection{Rule-based Methods}
Rule-based methods require to mine logical rules by learning frequent reasoning paths in KGs. Since the logical rules are composed of relations and independent of entities, rule-based methods are inherently well-suited for addressing inductive KGC. Neural-LP \cite{NeuralLP} proposed a framework to perform reasoning using logical rules in a differentiable manner, which uses matrix multiplication to mimic the inference process of logical rules. DRUM \cite{DRUM} developed an end-to-end differentiable rule mining model to mine logical rules and learn the corresponding confidence scores. RuleN \cite{RuleN} uses a statistic method to mine rules and determine their confidence. However, the performance of such methods is dependent on coverage and confidence of the mined logical rules, limiting their expressive ability.

\subsubsection{GNN-based Methods}
GNN-based inductive KGC methods have emerged as a dominant approach in recent years. GraIL \cite{GraIL} is one of the earlier works that employs GNNs for inductive reasoning. It extracts enclosing subgraphs for candidate triples and performs GNNs on these subgraphs. Thus, the link prediction task is converted to a subgraph classification task. CoMPILE \cite{CoMPILE} develops a stronger message passing method to improve GraIL, allowing sufficient information flow between entities and relations. TACT \cite{TACT} designs a relational correlation graph to learn topological structure of relations.  NodePiece \cite{NodePiece} utilizes relations as anchors to learn entity-independent embedding for nodes in KGs. MorsE \cite{MorsE} uses relation based entity initializer to learn entity-independent embebedding for each entity and uses GNN to enhance entity embedding by aggregating neighbor structure information. Meta-IKG \cite{MetaiKG} utilizes meta gradients to learn transferable patterns in local subgraphs. ConGLR \cite{ConGLR} extracts an enclosing subgraph and a context graph, then applies two GNNs to learn the structure of these two graphs. DEKG-ILP \cite{DEKGILP} uses relation-specific features and a contrastive learning method to predict bridging links between known entity and unseen entity.  VMCL \cite{VMCL} proposed a graph-guided variational autoencoder to enrich the entity features. RASC \cite{RASC} constructs a relational correlation subgraph and relational path subgraph to learn entity-independent node embeddings. LCILP \cite{LCILP} employs personalized PageRank to select important nodes in subgraphs. SNRI \cite{SNRI} enhances the entity embedding by using neighboring relational feature and neighboring relational path for sparse subgraph. SASR \cite{SASR} captures node structure information by counting the occurrences of specific substructures.  INGRAM \cite{InGram}, RMPI \cite{RMPI} and RAILD \cite{RAILD} are designed to address the inductive scenarios that both relations and entities are new in the emerging KGs. These GNN-based methods have achieved great performance gains due to their capability to capture local structure information. Nevertheless, most of these previous methods are based on enclosing subgraphs, which requires to extract a specific subgraph for each candidate triple,  resulting in inefficient inductive reasoning. In response to this issue, QAAR \cite{QAAR}  proposed a novel method based on opening subgraphs where all candidates for a query are collected in a single opening subgraph. The center node of the subgraph and its neighbors in one hop are used as anchors to learn entity-independent structure features. Despite its efficiency and effectiveness, the preliminary QAAR model still limits in learning sufficient structure features for entities far way from the center node. Therefore, we extend the previous QAAR \cite{QAAR} model to learn global-local anchor representation in a collaborative way. Specifically, our major extensions lie in the following two aspects: (1) a global anchor representation learning method is introduced to acquire entity-independent global features; (2) a global-local graph neural network is employed to incorporate both local and global features for effective inductive reasoning.

\section{Preliminaries}
\label{sec_pre}
{\bf Transductive knowledge graph completion (KGC).}
A KG is denoted as $\mathcal{G}=(\mathcal{E},\mathcal{R},\mathcal{T})$, where $\mathcal{E}$ denotes the entity set, $\mathcal{R}$ denotes the relation set, and $\mathcal{T}=\{(h,r,t)\} \subseteq \mathcal{E}\times \mathcal{R}\times \mathcal{E}$ denotes the triple set. Conventional transductive KGC methods learn the entity embeddings $\mathbf{E}\in \mathbb{R}^{|\mathcal{E}|\times dim}$ and the relation embeddings $\mathbf{R} \in \mathbb{R}^{|\mathcal{R}|\times dim}$, with the assumption that all the entities and relations in the test phase are seen in the training phase. A scoring function $f(h,r,t)$ is defined to predict the likelihood of a given triple $(h,r,t)$. Thus, the missing entities for a query $(h,r,?)$ or $(?,r,t)$ can be inferred by ranking candidates according to the predicted scores.

{\bf Inductive knowledge graph completion (KGC).}
Different from transductive KGC, inductive KGC aims to infer missing links between entities that are not observed during training. Formally, we define a training KG $\mathcal{G}_{train}=(\mathcal{E}_{train},\mathcal{R},\mathcal{T}_{train})$, where $\mathcal{E}_{train}$ denotes the training entity set and $\mathcal{T}_{train}=\{(h,r,t)\}\subseteq \mathcal{E}_{train}\times \mathcal{R}\times \mathcal{E}_{train}$ denotes a set of training triples. And let $\mathcal{G}_{test}=(\mathcal{E}_{test},\mathcal{R},\mathcal{T}_{test})$ represent a test graph, where $\mathcal{E}_{test}$ denotes a set of entities without overlap with the training entity set $\mathcal{E}_{train}$, namely $\mathcal{E}_{train}\bigcap \mathcal{E}_{test}=\emptyset$. The purpose of inductive KGC is to learn an entity-independent scoring function $f(h,r,t)$ on the training graph and then apply it to the test graph during the inference phrase. Thus, $f(h,r,t)$ can measure the likelihood for a given triple $(h,r,t)\in \mathcal{T}_{test}$ with unseen entities.

{\bf Enclosing subgraph.} Given a triple $(h,r,t)$, an enclosing subgraph \cite{GraIL} is a subgraph around the triple $(h,r,t)$, which is constructed by collecting all the paths between the head entity $h$ and the tail entity $t$. Thus, each candidate triple has a specific enclosing subgraph. 

{\bf Opening subgraph. } Different from the enclosing subgraph, we define the opening subgraph as a subgraph surrounding the query entity $h$ for a query $(h,r,?)$, which is induced by all the neighboring nodes of query entity $h$ within $k$-hop distance. Thus, all candidate answer entities can be covered in one opening subgraph and the entity $h$ is defined as the center node in the opening subgraph.

\begin{figure*}[t]
\centering
\includegraphics[width=0.92\textwidth]{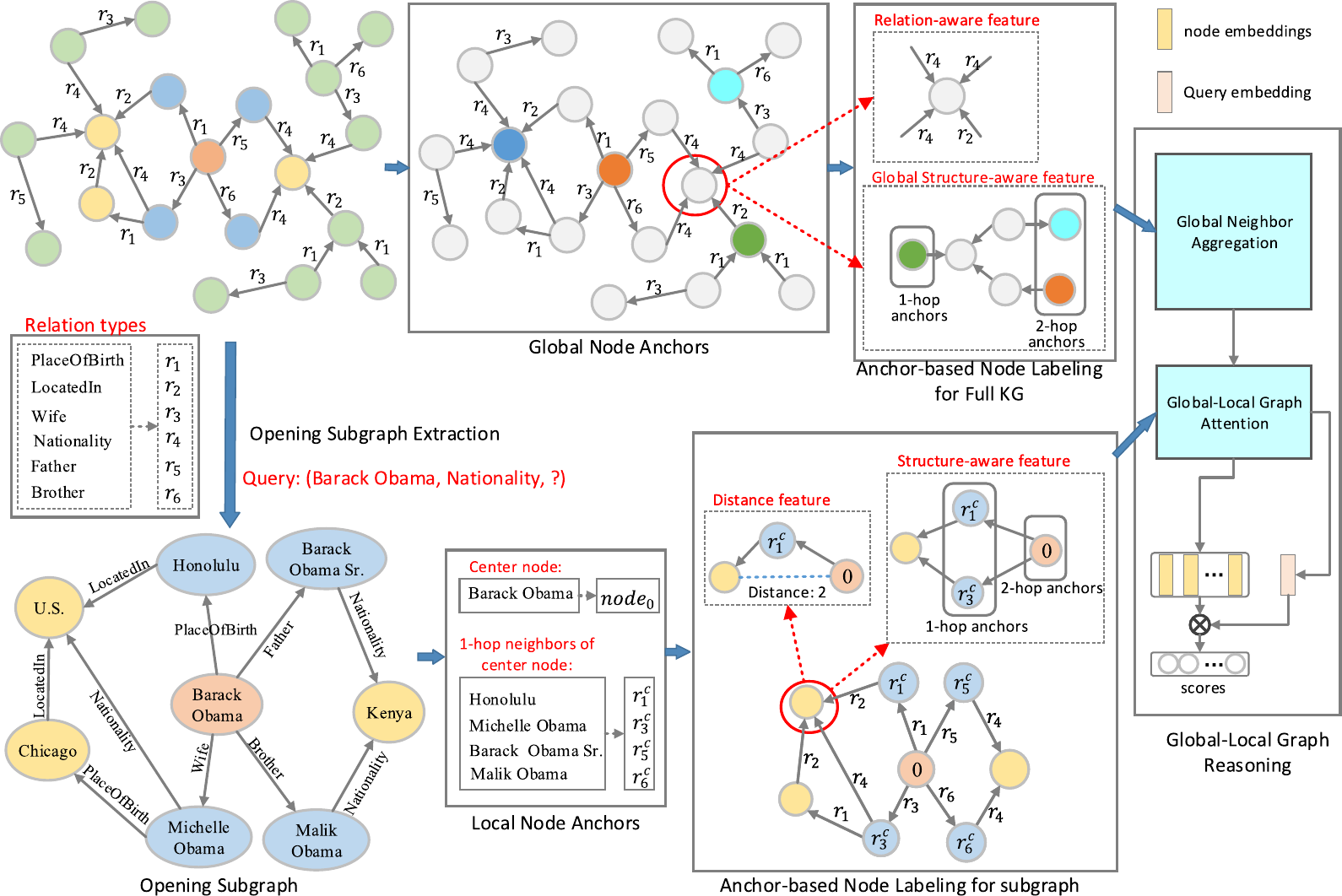}
\caption{The overview of the proposed GLAR model.} \label{fig_model}
\end{figure*}

\section{Our Approach}
\label{sec_method}
For inductive KGC, the model should be able to train on observed KGs while perform inferring on new KGs with unseen entities. Therefore, conventional transductive methods, which learn a specific embedding for each entity, cannot be applied to inductive setting. Instead, it is critical to learn entity-independent features for the nodes in a new KG. To address this challenge, we explore how to take full advantage of the structure information to learn entity-independent embeddings for the nodes in a given KG. 

We propose a novel global-local anchor representation (GLAR) for inductive KGC, which learns entity-independent node embeddings using both global and local structure information. Specifically, the proposed GLAR consists of a local anchor representation learning, a global anchor representation learning and a global-local graph reasoning module. The overview of the proposed GLAR model is illustrated in Figure~\ref{fig_model}. In the local anchor representation learning module, we firstly extract an opening subgraph around the entity $h$ in the given query $(h,r,?)$. Then, we define some local anchors, thus the nodes in the subgraph can be labeled based on these local anchors. In the global anchor representation learning module, we use a clustering method to select global anchors based on entity-independent relational features. Then, we label each node based on these global anchors.  Finally, a global-local graph reasoning network is applied to incorporate global and local node features and predict missing entities.

\subsection{Local Anchor Representation Learning}
\subsubsection{Opening Subgraph Extraction}
Intuitively, the multi-hop subgraph surrounding a given query $(h,r,?)$ contains various reasoning paths between the query entity $h$ and the answer entity, which can provide rich clues to predict missing answer entity of the query. Previous methods, such as GraIL \cite{GraIL} and its following works (e.g., TACT \cite{TACT}, ConGLR \cite{ConGLR} and CoMPILE \cite{CoMPILE}), construct a specific enclosing subgraph by extracting multi-hop paths from query entity $h$ and each candidate answer entity $t$ for every candidate triple $(h,r,t)$. However, these methods will construct multiple subgraphs for all candidate triples, which is inefficient when the number of candidates is large. Unlike previous methods, we extract one opening subgraph that can be shared by all candidate entities for the query $(h,r,?)$. Formally, the opening subgraph $\mathcal{G}_k(h)$ is constructed by extracting the neighboring nodes within $k$-hop surrounding the query entity $h$.

\subsubsection{Local Anchor Selection}
\label{sec_anchor_construction}
For inductive KGC, we need to learn rich node features that can be transferred to different opening subgraphs. For this purpose, we select some local anchors following two criteria: (1) the local anchors are independent of any specific entities and can be generalized to different subgraphs;  (2) the local anchors are desired to have the ability of capturing rich structure features. Based on these two criteria, we design some entity-independent local anchors in  the opening subgraph $\mathcal{G}_k(h)$, encompassing the center node and its immediate one-hop neighbors.

{\bf Center node}. In this study, the opening subgraph $\mathcal{G}_k(h)$ is constructed around the query entity $h$. Therefore, the query entity $h$ is viewed as the center node of the opening subgraph $\mathcal{G}_k(h)$. Since each opening subgraph has a unique center node, the center node can serve as a reference point to learn the graph structure. Thus, we define the center node as an anchor and represent it as a special symbol $node_0$.

{\bf One-hop neighbors regarding the center node}. Additionally, we designate the one-hop neighbors regarding the center node as anchor nodes to enhance the learning of the local subgraph structure. In inductive KGC task, the one-hop neighbors of the center node cannot be represented as entity-specific embeddings.  To tackle this issue, we utilize the relationship between the center node and each neighboring node to represent these neighbors. In particular, when considering a neighboring entity $t$ linked to the central node $h$ through a triple $(h,r_i,t)$, we can represent the neighboring entity $t$ as $r_i^c$, denoting a node connected to the center node through relation $r_i$.
Considering the triple ({\em Barack Obama}, {\em PlaceOfBirth}, {\em Honolulu}) as an example, we can represent the entity {\em Honolulu} as ``the PlaceofBirth of Barack Obama". In this manner, we can represent all neighboring nodes of the center node via the combination the center node and the relation, thus resulting in a set of one-hop neighborhood  anchors $D=\{r_i^c|(h,r_i,t) \in \mathcal{T}_k(h)\}$, where $\mathcal{T}_k(h)$ is the triple set in $\mathcal{G}_k(h)$.

We define the center node and its one-hop neighboring nodes as a set of local anchors in the opening subgraph, which is denoted as $Anc=\{node_0\}\cup D$. Due to the entity-independent characteristics of these anchors, they can serve as transferable meta-knowledge to narrow the gap between the training KG  and the test KG.

\subsubsection{Anchor-based Node Labeling}
\label{sec_node_labeling}
In this subsection, we explain the process of labeling each node in the opening subgraph based on the local anchors $Anc$. In particular, we define two kinds of local features, namely the local structure features and the distance features.

{\bf Local structure features}. In this study, we design a structure-aware labeling method to capture structure features of each node by utilizing the neighboring relationship between each node and the defined local anchors.  Formally, we denote $\mathcal{A}_j(n)=\{a_i|a_i \in \mathcal{N}_j(n)\bigcap Anc \}$ as the neighboring anchor nodes of node $n$ at the $j$-th hop, where $\mathcal{N}_j(n)$ represents the $j$-th hop neighbors of node $n$.  Then, the local structure feature of node $n$ at the $j$-th hop can be computed as:
\begin{equation}
    \mathbf{v}_n^j=\sum_{a_i \in \mathcal{A}_j(n)} \mbox{onehot}({a}_i)
\end{equation}
where $\mbox{onehot}(a_i)$ is a one-hot vector with a size of $|Anc|$ where the $i$-th elements is $1$ and other element is $0$. By collecting neighboring anchor nodes from various hops, a set of local structure features can be acquired, denoted as $\{\mathbf{v}_n^0,\mathbf{v}_n^1,\cdots,\mathbf{v}_n^J\}$, where $J$ is the maximum hop of the structure-aware features.

{\bf Distance features}. In addition, the distances between the center node and other nodes are also an important feature to learn the graph structure. Consequently, we utilize distance features to enhance the node embeddings. Concretely, we start from the center node to traverse all nodes through breadth-first search to obtain the distances from the central node to other nodes. Then, the vector of distance feature is denoted as:
\begin{equation}
    \mathbf{v}_n^d = \mbox{onehot}(d_n)
\end{equation}
where $d_n$ is the distance between the center node and node $n$, $\mbox{onehot}(d_n)$ is a one-hot vector of the distance $d_n$.

Finally, we combine the local structure feature and distance feature to generate the initial local node feature:
\begin{equation}
    \mathbf{e}^0_n = [\mathbf{v}_n^0||\mathbf{v}_n^1||\cdots||\mathbf{v}_n^J||\mathbf{v}_n^d] \mathbf{W}^0+\mathbf{b}^0
\end{equation}
where $\mathbf{W}^0$ and $\mathbf{b}^0$ are the trainable parameters.

\subsection{Global Anchor Representation Learning}
By learning the local anchors, we can obtain the node features according to the local structure of the opening subgraph. Unfortunately, the selected local anchors tend to concentrate primarily around the center node, resulting in an imbalance of features for different nodes. Nodes near the center of the subgraph will learn richer structure features, whereas nodes far from the center of the subgraph cannot learn effective structure features, especially for nodes outside of the subgraph.  To alleviate this problem, we propose a global anchor representation learning method to select representative anchor nodes in the global KG.

\subsubsection{Entity Independent Global Anchor Selection}
To compensate for the shortcomings of local structure features, we propose to select global anchors following two criteria: (1) global anchors are entity-independent and can be transferred from training KG to test KG;  (2) global anchors should be able to provide structure information for as many nodes as possible. To this end,  we design a clustering-based method to select global anchors. Since entity-independent features are required for inductive KGC, we use neighboring relations as node features to select global anchors.  Formally, we define the relational feature of the node $n$ as:
\begin{equation}
    \mathbf{v}^r_n = \sum_{r\in \mathcal{R}(n)} \mbox{onehot}(r)
    \label{eq_relation_feat}
\end{equation}
where $\mathcal{R}(n)$ is a set of neighboring relations of node $n$, and $\mbox{onehot(r)}$ is a one-hot vector of the relation $r$. Thus, all nodes in the training KG $\mathcal{G}_{train}$ can be represented by the relational feature and the node feature matrix can be obtained by stacking all these node features:
\begin{equation}
    \mathbf{H}_{\mathcal{G}_{train}} = \mathop{||}\limits_{n\in \mathcal{E}_{train}} \mathbf{v}^r_n
\end{equation}
Similarly, we can obtain the node feature matrix for testing KG  $\mathcal{G}_{test}$, which is defined as $\mathbf{H}_{\mathcal{G}_{test}}$.

Then, we train a clustering model over the relational features of training KG $\mathbf{H}_{\mathcal{G}_{train}}$, where the cluster number is set as $m$. Formally, the clustering model can be defined as:
\begin{equation}
    \mathcal{C}=\mbox{clustering}(\mathbf{H}_{\mathcal{G}_{train}},m)
\end{equation}
where $\mathcal{C}=\{\mathbf{c}_1,\mathbf{c}_2,\cdots,\mathbf{c}_m\}$ is a set of cluster centroids for the clusters and $c_i$ is the feature vector of the $i$-th centeroid.  

Finally, we can select global anchor nodes based on these cluster centroids. Specifically, we assign a cluster label to each node based on its distance to the cluster center and partition nodes into different groups according to their cluster labels. A representative node with the largest degree is selected as global anchor node for each cluster:
\begin{equation}
    a_{c_i}=\mbox{TopDegree}(\mathcal{E}_{c_i})
\end{equation}
where $\mathcal{E}_{c_i}$ denotes the node set of the cluster $c_i$ in training KG or testing KG, $\mbox{TopDegree}(\cdot)$ denotes a function to select global anchor node with the largest degree. Thus, we can obtain a set of global anchor nodes $Anc_{global}=\{a_{c_0},a_{c_1}, \cdots, a_{c_m}\}$.

\subsubsection{Node Labeling with Global Anchors}
For the nodes in the full KG (e.g., $\mathcal{G}_{train}$ or $\mathcal{G}_{test}$ ), we use two kinds of entity-independent features to label each node, including relational features and global structure features.

{\bf Relational features for nodes in full KG.} Given a node $n$ in the full KG $\mathcal{G}_{train}$ (or $\mathcal{G}_{test}$), we use its neighboring relational feature $\mathbf{v}^r_n$ defined in Equation~\ref{eq_relation_feat} as one of entity-independent features to learn the node representation.  

{\bf Global structure features for nodes in full KG.} Simultaneously, we can learn global structure features for nodes in full KG by capturing neighboring global anchors for each node. Formally, let $\mathcal{A}^g_j=\{a_c|a_c \in \mathcal{N}_j(n)\bigcap Anc_{global}\}$ denote the $j$-th hop neighboring global anchors.  We can obtain the $j$-th hop global structure feature for node $n$:
\begin{equation}
    \mathbf{v}^j_{gn}= \sum_{a_c \in \mathcal{A}^g_j} \mbox{onehot}(a_c)
\end{equation}
where $\mbox{onehot}(a_c)$ is a $m$ dimensional one-hop vector for global anchor $a_c$. Similar to the local structure feature, we can also capture global structure features from different hops and obtain multi-hop global structure feature:
\begin{equation}
    \mathbf{v}_{gn}=\mathop{||}\limits_{j=0}^J \mathbf{v}^j_{gn} 
\end{equation} 

Finally, the global node feature in full KG is computed by merging these two kinds of features:
\begin{equation}
    \mathbf{e}^0_{gn} = \mathbf{v}^r_n \mathbf{W}^r + \mathbf{v}_{gn} \mathbf{W}^g
\end{equation}
where $\mathbf{W}^r$ and $\mathbf{W}^g$ are trainable parameters.

\subsection{Global-Local Graph Reasoning}
In the above sections, we have introduced how to obtain entity-independent node features for opening subgraphs and full KGs. Subsequently,  we propose a global-local graph reasoning method to perform inductive reasoning on KGs by
collaboratively propagating both local and global information. In the global-local reasoning model, we firstly apply a graph convolutional network to aggregate global neighborhood information on the full KG, then use a global-local graph attention mechanism to learn global-local node embeddings by incorporating global neighborhood features with local neighborhood features. Finally, global-local node embeddings are used to compute the scores for candidate entities.

Specifically, given the query $(h,r,?)$, we firstly aggregate global neighborhood information for each node $n$ by performing a graph convolutional network on the full KG:
\begin{equation}
    \mathbf{e}^{l+1}_{gn} =\frac{1}{|\mathcal{N}_{g}(n)|} \sum_{(r_p,p)\in \mathcal{N}_{g}(n)} [\mathbf{e}^{l}_{gp}||\mathbf{r}_p] \mathbf{W}^{l}_{g}+\mathbf{b}^l_{g}
\end{equation}
where $\mathcal{N}_{g}(n)$ represents the collection of global neighbors for node $n$, which can be formulated as $\mathcal{N}_{g}(n)=\{(r_p,p)|(h,r_p,p)\in \mathcal{T}_{train}\}$ during training and $\mathcal{N}_{g}(n)=\{(r_p,p)|(h,r_p,p)\in \mathcal{T}_{test}\}$ during testing, $\mathbf{e}^{l}_{gp}$ denotes the global node embedding for the neighboring node $p$ in the $l$-th layer.

Then, we collaboratively aggregate the global-local neighborhood features in the opening subgraph $\mathcal{G}_k(h)$ for each node $n$ by integrating global node embeddings with local node embeddings, which is defined as:
\begin{equation}
    \mathbf{e}'_n = \sum_{(r_z,z) \in \mathcal{N}(n)} \alpha^l_{z} [\mathbf{e}_z^l||\mathbf{e}^{l}_{gz}||\mathbf{r}_z]\mathbf{W}^l+\mathbf{b}^l
\end{equation}
where $\mathcal{N}(n)=\{(r_z,z)|(n,r_z,z) \in \mathcal{T}_k(h)\}$ is the collection of neighboring nodes for node $n$, $\mathcal{T}_k(h)$ represents the set of triples in $\mathcal{G}_k(h)$, $\mathbf{e}^{l}_{z}$ and $\mathbf{e}^{l}_{gz}$ denote the local and global node embedding for the neighboring node $z$,  and $\alpha^l_z$ denotes the attention score for the neighbor $(r_z,z)$ which is calculated as:
\begin{equation}
    \alpha^l_z = \sigma(\mathbf{r}_a^T([\mathbf{e}_n^l||\mathbf{e}^{l}_{gz}||\mathbf{r}_{z}||\mathbf{e}_z^l]\mathbf{W}^l_a+\mathbf{b}_a^l))
\end{equation}
where  $\mathbf{r}_a$, $\mathbf{W}^l_a$ and $\mathbf{b}_a^l$ are learnable parameters, $\sigma$ represents the sigmoid function. 

In order to mitigate the issue of gradient vanishing, we employ a gated mechanism that combines both the node embeddings and their global-local neighboring features.
\begin{equation}
    \mathbf{e}^{l+1}_n = \beta  \times [\mathbf{e}^{l}_n||\mathbf{e}^{l}_{gn}] \mathbf{W}^l_{\beta} + (1-\beta)\times \mathbf{e}'_n
\end{equation}
where $\beta$ represents the gate used to regulate the weighting between the neighborhood feature $\mathbf{e}'_n$ and the node features $\mathbf{e}^{l}_n$ and $\mathbf{e}^{l}_{gn}$, which is defined as:
\begin{equation}
\beta=\sigma([\mathbf{e}^{l}_n||\mathbf{e}^{l}_{gn}||\mathbf{e}'_n]\mathbf{W}_{\beta'}^l+\mathbf{b}_{\beta'}^l)
\end{equation}
Note that for the node which is not in the opening subgraph $\mathcal{G}_k(h)$, its local node embedding $\mathbf{e}^l_n$ is set to zero vector since there is no local information available for it, while the global node embedding is applied to provide structure information for it. In this way, global and local node embeddings are collaboratively used to capture rich structure features for all nodes. 

We stack $L$ layers of global-local graph neural networks to capture multi-hop graph structure. And the final node features are obtained via a concatenation of node vectors in all layers followed with a linear feed-forward layer:
\begin{equation}
    \mathbf{e}_n=[\mathbf{e}^0_n||\mathbf{e}_n^1||\cdots||\mathbf{e}_n^L] \mathbf{W}_o + \mathbf{b}_o
\end{equation}

Sequentially, the query embedding of  $(h,r,?)$ is computed as:
\begin{equation}
    \mathbf{e}_{query}=[\mathbf{e}_{\mathcal{G}}||\mathbf{e}_h||\mathbf{r}]\mathbf{W}_q + \mathbf{b}_q
\end{equation}
where $\mathbf{e}_h$ is the embedding for entity $h$, $\mathbf{r}$ is the embedding for relation $r$ in the query, $\mathbf{e}_{\mathcal{G}}$ is the embedding of the subgraph $\mathcal{G}_k(h)$, which is computed as:
\begin{equation}
    \mathbf{e}_{\mathcal{G}}=\frac{1}{|\mathcal{E}_k|}\sum_{n \in \mathcal{E}_k} \mathbf{e}_n
\end{equation}

Finally, we can calculate the score for each candidate answer entity $t$ by taking the inner product of candidate answer entity embedding $\mathbf{e}_{t}$ and the query embedding $\mathbf{e}_{query}$:
\begin{equation}
    f(h,r,t)=\sigma(\mathbf{e}^T_{query}\mathbf{e}_{t})
\end{equation}
where $\mathbf{e}_t$ is the embedding of the candidate entity $t$.

To train the model, we employ a cross-entropy loss function to encourage the model to produce higher scores for positive samples and lower scores for negative samples:
\begin{equation}
    \mathcal{L}= - \sum_{(h,r,t)\in \mathcal{G}(h)} \mbox{log}f(h,r,t) + \mbox{log}(1-f(h,r,t')) 
\end{equation}
where $t'$ is a negative entity for the query $(h,r,?)$.

When dealing with the query $(?, r, t)$, we apply an inverse transformation by using the form $(t, r^{-1}, ?)$, which effectively converts it into a query with a missing tail entity.

\begin{algorithm}[tb]
\small
\caption{Structure feature extraction.}
\label{alg:algorithm}
\textbf{Input}: Subgraph $\mathcal{G}_k(h)$ and anchor nodes $Anc$. \\
\textbf{Output}: Structure-aware anchor set $\mathcal{A}_j(n)$ for each node $n$ in subgrah $\mathcal{G}_k(h)$.
\begin{algorithmic}[1] 
\STATE Initialize an empty set $\mathcal{A}_j(n)$ for each node $n$.
\FOR{$a_i \in Anc$}{
\STATE Initialize a queue $S=\{(a_i,0)\}$.
\WHILE{$S$ is not empty}{
\STATE Take an element $(n,j)$ out from the queue $S$, where $j$ denotes the distance from node $n$ to the anchor node $a_i$.
\IF{$j\leq J$} {
\STATE Add the anchor node $a_i$ to $\mathcal{A}_j(n)$.
\FOR{each node $z\in \mathcal{N}(n)$}{
\IF{$z$ has not been visited}{
\STATE Add  $(z,j+1)$ into queue $S$.
}
\ENDIF
}
\ENDFOR
}
\ENDIF
}
\ENDWHILE
}
\ENDFOR
\end{algorithmic}
\end{algorithm}

\begin{table}
\caption{Statistics of WN18RR-ind, FB15k237-ind and NELL995-ind datasets. "\#R", "\#E" and "\#T" are the numbers of relations, entities and triples. }
\centering
\resizebox{\columnwidth}{!}{
\begin{tabular}{cccccccccccc}
\hline
& &\multicolumn{3}{c}{WN18RR-ind}  &\multicolumn{3}{c}{FB15k237-ind}  &\multicolumn{3}{c}{NELL995-ind}  \\
& &\#R &\#E &\#T &\#R &\#E &\#T &\#R &\#E &\#T\\
\hline
\multirow{2}{*}{v1} &train &9 &2746 &6678 &183 &2000 &5226 &14 &10915 &5540 \\
&test &9 &922 &1991 &146 &1500 &2404 &14 &225 &1034\\
\hline
\multirow{2}{*}{v2} &train &10 &6954 &18968 &203 &3000 &12085 &88 &2564 &10109\\
&test &10 &2923 &4863 &176 &2000 &5092 &79 &4937 &5521\\
\hline
\multirow{2}{*}{v3} &train &11 &12078 &32150 &218 &4000 &22394 &142 &4647 &20117 \\
&test &11 &5084 &7470 &187 &3000 &9137 &122 &4921 &9668\\
\hline
\multirow{2}{*}{v4} &train &9  &3861 &9842 &222 &5000 &33916 &77 &2092 &9289\\
&test &9 &7208 &15157 &204 &3500 &14554 &61 &3294 &8520 \\
\hline
\end{tabular}
}
\label{tab_datasets}
\end{table}

\subsection{Complexity Analysis}
To obtain the structure-aware anchor set $\mathcal{A}_j(n)$ (or $\mathcal{A}^g_j(n)$), one simple way is to visit each node and then use breadth-first search\footnote{https://en.wikipedia.org/wiki/Breadth-first\_search} method to extract its neighbors from different hops for every node. However, the overall time complexity is $O(|\mathcal{E}_k|^2)$, where $\mathcal{E}_k$ is the set of entities in subgraph $\mathcal{G}_k(h)$. It is time-consuming since the number of entities in $\mathcal{E}_k$ is usually very large. To alleviate this issue, we develop an efficient structure-aware feature extraction algorithm shown in Algorithm~\ref{alg:algorithm}. Instead of visiting all the nodes in the subgraph, we only visit the anchor nodes in the subgraph. The time complexity to extract local structure features is reduced to $O(|Anc|\times|\mathcal{E}_k|)$, where $|Anc| << |\mathcal{E}_k|$. Similarly, the time complexity for global structure features is $O(|Anc_{global}|\times |\mathcal{E}|)$.

To obtain the distance features, we need visit all the nodes in the opening subgraph, therefore the time complexity of extracting distance features is $O(|\mathcal{E}_k|)$. And the time complexity of relational features is $O(|\mathcal{R}|\times |\mathcal{E}|)$. In the global-local reasoning module, we aggregate the neighbors of each entity using global-local graph neural networks, thus the time complexity of the aggregation operation is $O(|\mathcal{E}|\times |\mathcal{N}|+|\mathcal{E}_k|\times |\mathcal{N}|)$, where $|\mathcal{N}|$ is the number of neighbors for each entity. Since $|\mathcal{E}_k|\leq |\mathcal{E}|$, we can omit $|\mathcal{E}_k|$ and represent the overall time complexity as $O(|\mathcal{E}|\times Q)$, where $Q=2|\mathcal{N}|+|\mathcal{R}|+|Anc|+|Anc_{global}|+1$.

It is important to note that all candidate answer entities reside within a single opening subgraph $\mathcal{G}_k(h)$. Consequently, we can calculate scores for all candidates through a single round of reasoning within the subgraph $\mathcal{G}_k(h)$. This differs from previous enclosing subgraph based methods that necessitate the extraction and processing of a distinct subgraph for every candidate triple. Specifically, for a query with $P$ candidates, previous enclosing subgraph based methods will perform reasoning for $P$ times with the time complexity of $O(|\mathcal{E}|\times|\mathcal{N}|\times P)$, while our method will only perform reasoning for one time with the time complexity $O(|\mathcal{E}|\times Q)$. Since the number of candidates is always large for KGC task, $O(|\mathcal{E}|\times|\mathcal{N}|\times P)$ is far greater than $O(|\mathcal{E}|\times Q)$. Therefore, our opening subgraph based method is more efficient to perform reasoning on the inductive KGC task than the previous enclosing subgraph based methods.

\begin{table*}
\centering
\caption{Hits@10 performance over the inductive KGC benchmarks WN18RR-ind, FB15k237-ind and NELL995-ind. The results with $^\dagger$ are taken from \cite{ConGLR}.  “Avg.” denotes the average performance of the four versions. }
\setlength{\aboverulesep}{0pt}
\setlength{\belowrulesep}{0pt}
\begin{tabular}{c|ccccc|ccccc|ccccc}
\toprule
\multirow{2}{*}{Models}&\multicolumn{5}{c|}{WN18RR-ind}  &\multicolumn{5}{c}{FB15k237-ind} &\multicolumn{5}{c}{NELL995-ind}  \\
 &v1 &v2 &v3 &v4 &Avg. &v1 &v2 &v3 &v4 &Avg. &v1 &v2 &v3 &v4 &Avg.\\
 \hline
 Neural-LP$^\dagger$&74.37 &68.93 &46.18 &67.13 &64.15 &52.92 &58.94 &52.90 &55.88&55.16  &40.78 &78.73 &82.71 &80.58 &70.70\\
 DRUM$^\dagger$&74.37 &68.93 &46.18 &67.13 &64.15 &52.92 &58.73 &52.90 &55.88 &55.10 &19.42 &78.55 &82.71 &80.58 &65.31\\
 RuleN$^\dagger$&80.85 &78.23 &53.39 &71.59 &71.01 &49.76 &77.82 &87.69 &85.60 &75.21 &53.50 &81.75 &77.26 &61.35 &68.46\\
 \hline
 \hline
Meta-iKG&- &- &- &- &- &66.96 &74.08 &71.89 &72.28 &71.30 &64.20 &77.91 &77.41 &73.12 &73.16\\
MorsE &84.14 &81.50 &70.92 &79.61 &79.04 &83.17 &{95.67} &{95.69} &{95.89} &{92.61} &65.20 &80.70 &87.67 &53.44 &71.75\\
GraIL$^\dagger$&82.45 &78.68 &58.43 &73.41 &73.24 &64.15 &81.80 &82.83 &89.29 &79.51 &59.50 &93.25 &91.41 &73.19 &79.33\\
CoMPILE$^\dagger$&83.60 &79.82 &60.69 &75.49 &74.90 &67.64 &82.98 &84.67 &87.44 &80.68 &58.38 &93.87 &92.77 &75.19 &80.05\\
TACT$^\dagger$ &84.04 &81.63 &67.97 &76.56 &77.55 &65.76 &83.56 &85.20&88.69 &80.80 &79.80 &88.91 &94.02 &73.78 &84.12 \\ 
SNRI & 87.23 &83.10 & 67.31 &83.32&80.24 &71.79 &86.5&89.59&89.39&84.32&- &- &- &- &- \\
LCILP&88.30 &84.12 &72.68 &79.46 &81.14 &70.97 &81.89 &84.33 &89.84 &81.75&52.40 &93.58 &92.15 &82.90 &80.25 \\
ConGLR$^\dagger$&{85.64} &{92.93} &70.74 &\underline{92.90}&{85.55} &68.29 &85.98 &88.61 &89.31 &82.93 &81.07 &{94.92} &{94.36} &81.61 &87.99\\
NodePiece&83.00 &{88.60} &{78.50} &80.70 &82.70 &{87.30} &93.90 &94.40 &{94.90} &92.35 &{89.00} &90.10 &93.60 &{89.30} &{90.50} \\
NBFNet &\underline{94.80} & 90.50 &\underline{89.30} &89.00 &{90.90} &83.40 &94.90 &95.10&96.00 &92.35 &- &- &- &- &- \\
REST&\bf{96.28} &\underline{94.56} &79.50 &\bf{94.19} &\underline{91.13} &75.12 &91.21 &93.06 &96.06 &88.86 &88.00 &94.96 &\bf{96.79}&\underline{92.61} &\underline{93.09} \\
QAAR &{88.82} &86.87 &{88.18} &{89.15} &{88.25} &\underline{90.48} &\underline{95.81} &\underline{95.87} &\underline{96.21}&\underline{94.59} &\underline{89.50} &\underline{96.32} &{94.56} &{91.15}&{92.88}\\
GLAR &{93.68} &\bf{94.74} &\bf{93.32} &{92.44} &\bf{93.55} &\bf{91.34} &\bf{96.59} &\bf{95.95} &\bf{96.37} &\bf{95.06} &\bf{91.5} &\bf{97.26} &\underline{95.61} &\bf{93.16} &\bf{94.38}\\
\bottomrule
\end{tabular}

\label{tab_res_hit10}
\end{table*}

\begin{table*}[!t]
\caption{AUC-PR performance over the inductive benchmarks WN18RR-ind, FB15k237-ind, NELL995-ind. }
\centering
\setlength{\aboverulesep}{0pt}
\setlength{\belowrulesep}{0pt}
\begin{tabular}{c|ccccc|ccccc|ccccc}
\toprule
\multirow{2}{*}{Models}&\multicolumn{5}{c}{WN18RR-ind} &\multicolumn{5}{c}{FB15k237-ind} &\multicolumn{5}{c}{NELL995-ind}    \\
 &v1 &v2 &v3 &v4 &Avg. &v1 &v2 &v3 &v4 &Avg. &v1 &v2 &v3 &v4 &Avg.\\
 \hline
 Neural-LP$^\dagger$&86.02 &83.78 &62.90 &82.06 &78.69 & 69.64&76.55&73.95&75.74&73.97 &64.66 &83.61 &87.58 &85.69 &80.38 \\
 DRUM$^\dagger$&86.02 &84.05 &63.20 &82.06 &78.83 &69.71 &76.44 &74.03 &76.2 &74.09 &59.86 &83.99 &87.71 &85.94 &79.37\\
 RuleN$^\dagger$&90.26 &89.01 &76.46 &85.75 &85.37 &75.24 &88.70 &91.24 &91.79 &86.74 &84.99 &88.4 &87.2 &80.52 &85.27 \\
 \hline
 \hline
 Meta-iKG&-&-&-&-&- &81.10 &84.26 &84.57 &83.70 &83.41 &72.50 &85.97 &84.05 &81.24 &80.94\\
GraIL$^\dagger$&94.32 &94.18 &85.8 &92.72 &91.75 &84.69 &90.57 &91.68 &94.46 &90.35 &86.05 &92.62 &93.34 &87.50 &89.87 \\
CoMPILE$^\dagger$&{98.23} &{99.56} &93.60 &\underline{99.80} &{97.79} &85.50 &91.68 &93.12 &94.90 &91.30 &80.16 &{95.88} &96.08 &85.48 &89.40\\
TACT$^\dagger$ &95.43 &97.54 &87.65 &96.04 &94.16 &83.15 &{93.01} &92.1 &94.25 &90.62 &81.06 &93.12 &96.07 &85.75 &89.00\\
SNRI&99.10&\bf{99.92}&94.90&99.61&\underline{98.38}&86.69&91.77&91.22&93.37&90.76&-&-&-&-&- \\
LCILP&95.51&96.86&90.87&94.12&94.34&85.64&91.15&92.93&94.63&91.08&79.23&94.31&94.10&89.92&89.39\\
ConGLR$^\dagger$&\bf{99.58} &\underline{99.67} &{93.78} &\bf{99.88} &{98.22} &{85.68} &92.32 &{93.91} &{95.05} &{91.74} &{86.48} &95.22 &{96.16} &{88.46} &{91.58}\\
SASR &\underline{99.57} &99.00 &94.25 &98.56 &97.85 &92.21 &95.81 &\bf{97.83} &\bf{98.64}&96.12 &\underline{96.83}&97.56 &\underline{96.45} &\bf{97.24} &\underline{97.02} \\
QAAR &96.82 &94.15 &\underline{95.11} &96.57 &95.66 &{95.27} &\bf{97.71} &{97.52} &{97.47} &\underline{96.99} &{95.19} &\underline{97.64} &{96.19} &{94.35} &{95.84}\\
GLAR &98.68 &98.85 &\bf{97.19} &98.99 &\bf{98.42} &\bf{95.85} &\underline{97.49} &\underline{97.69} &\underline{97.75} &\bf{97.19} &\bf{97.11} &\bf{97.74} &\bf{97.12} &\underline{96.29} &\bf{97.06}\\
\bottomrule
\end{tabular}
\label{tab_res_auc}
\end{table*}

\section{Experiments}
\label{sec_result}
\subsection{Datasets}
In this study, we conduct experiments on three widely used benchmark datasets for inductive KGC, which are derived from WN18RR \cite{ConvE}, FB15k-237\cite{FB15k237} and NELL-995 \cite{DeepPath} by sampling disjoint subgraphs from the original KGs \cite{GraIL}. In this paper, these inductive datasets are referred to as WN18RR-ind, FB15k237-ind and NELL995-ind. These datasets follows a full-inductive setting where the train and test KG have no overlapping entities. Each inductive dataset has four versions with increasing sizes. Table~\ref{tab_datasets} summarizes the statistics of these datasets. 

\subsection{Evaluation Metrics}
Following previous works \cite{GraIL,CoMPILE,ConGLR},  we apply both classification and ranking metrics to measure the performance of the models. Concretely, we use AUC-PR (e.g., area under the precision-recall curve) to evaluate classification performance and Hits@10 to evaluate ranking performance, respectively. For the AUC-PR metric \cite{GraIL, TACT, ConGLR} , we randomly sample a negative sample for each positive triple by replacing the head or tail entity to ensure the  ratio of positive and negative samples is 1:1 and compute the AUC-PR based on the scores of positive and negative samples. To calculate hits@10, we follow previous works \cite{GraIL,TACT,ConGLR} to predict the missing tail (or head) entity for the query $(h,r,?)$ (or $(?,r,t)$) by ranking the candidate answer entities against other negative entities and computing the proportion of samples ranked in top 10.

\subsection{Experimental Setting}
Our proposed method is implemented using Pytorch and DGL \cite{DGL} framework. We conduct experiments on a single NVIDIA RTX A5000 GPU with 24GB memory. The clustering model is implemented using KMeans in scikit-learn \footnote{https://scikit-learn.org}. To train our model, the Adam \cite{Adam} optimizer is applied to learn trainable parameters in the model. In our experiments, we use a grid search method to select optimal hyper-parameters.  Specifically, we select the batch size from a range of  \{8,16,32\}, the learning rate from \{0.0001,0.0005,0.001,0.002\}, the number of global-local graph reasoning layers from \{1,2,3\}, the size $k$ of the opening subgraph  from \{3,4,5,6\}, the maximum hop of structure feature $J$ from a range of \{1,2,3\}, the number of global anchors from a range of \{50,100,150,200\}. Finally, the optimal hyper-parameters are as follows: the batch size is 16, the learning rate is 0.001, the number of global-local graph reasoning layer is 2, the size of subgraph is $k=6$, the number of global anchors is set to $m=100$, and the maximum hop of structure feature is $J=2$.

\subsection{Baselines}
To investigate the performance of the proposed GLAR model for inductive KGC, we compare our model with several strong baselines including rule-based methods and GNN-based methods. 

{\bf Rule-based methods} require to extract logical rules from KGs and perform reasoning based on these rules, including Neural-LP \cite{NeuralLP}, DRUM \cite{DRUM} and RuleN \cite{RuleN}. {\bf GNN-based methods} attempt to learn entity-independent features for KGs and use GNNs to learn the structure feature of KGs, including Meta-iKG\cite{MetaiKG}, MorsE \cite{MorsE}, GraIL \cite{GraIL}, CoMPILE \cite{CoMPILE}, TACT \cite{TACT}, SNRI \cite{SNRI}, LCILP \cite{LCILP}, ConGLR \cite{ConGLR}, NodePiece \cite{NodePiece}, NBFNet \cite{NBFNet}, REST \cite{REST} and QAAR \cite{QAAR}.

Since the text-based methods require additional entity and relation descriptions which are not available in the datasets, we do not compare with the text-based methods in our experiments.

\subsection{Experimental Results}
\subsubsection{Performance on Ranking Metric}
We compare the ranking performance of the proposed GLAR model with various strong baselines in Table~\ref{tab_res_hit10} where the results are measured using hits@10. From Table~\ref{tab_res_hit10}, we can observe several interesting discoveries:

\begin{itemize}
\item Compared to the rule-based methods, the GNN-based methods achieve considerable performance improvements on hits@10 metric across all these three inductive datasets. For example, the GraIL model outperforms RuleN by a margin of 2.23\% on WN18RR-ind, 4.3\% on FB15k237-ind and 10.87\% on NELL995-ind. And other SOTA GNN-based methods (e.g., CoMPILE, TACT, ConGLR and etc.) can further improve the performance on these inductive datasets. This is because that the performance of rule-based methods are heavily relied on the coverage and confidence of mined logical rules, making them inflexible to perform reasoning. Different from rule-based methods, the GNN-based models can infer missing links by capturing structure patterns via GNNs, which is more robust to perform reasoning.

\item The proposed GLAR model consistently outperforms most of SOTA rule-based and GNN-based models on hits@10 metric.  In particular, our GLAR model achieves the best hits@10 results on 9 testing sets among all 12 versions of the inductive datasets. And our GLAR model surpasses other SOTA models by large margins on the average hits@10 metrics of these three datasets. Specifically, our GLAR model outperforms the previous SOTA model REST by 2.42\% on WN18RR-ind, 6.4\% on FB15k237-ind and 1.29\% on NELL995-ind. This demonstrates the effectiveness of our GLAR model.

\item Compared to previous SOTA model ConGLR which is based on enclosing subgraph, our GLAR model gains 8\% improvement in term of average hits@10 on WN188RR-ind, 12.13\% improvement on FB15k237-ind and 6.39\% improvement on NELL995-ind. These experimental results demonstrate that our GLAR method, which is based on opening subgraphs, is superior to the enclosing subgraph based methods.

\item Compared to the previous work QAAR which also uses opening subgraphs, the proposed GLAR model also achieves better hits@10 results on all these three inductive datasets. Specifically, the GLAR model outperforms the QAAR model by a margin of 5.3\% in term of average hits@10 on WN188RR-ind, a margin of 0.47\% on FB15k237-ind and a margin of 1.5 \% on NELL995-ind. The reason is that the proposed GLAR model considers both global and local structure features in KGs to enhance the representation of entities, enabling the model to capture richer structure information.
\end{itemize}

\begin{table*}
\caption{Ablation study. Hits@10 results on the inductive KGC datasets. $\Delta$ represents the distinctions between the original GLAR model and its variant. }
\centering
\setlength{\aboverulesep}{0pt}
\setlength{\belowrulesep}{0pt}
\begin{tabular}{c|ccccc|ccccc|ccccc}
\toprule
\multirow{2}{*}{Models}&\multicolumn{5}{c|}{WN18RR-ind}  &\multicolumn{5}{c}{FB15k237-ind} &\multicolumn{5}{c}{NELL995-ind} \\
 &v1 &v2 &v3 &v4 &Avg. &v1 &v2 &v3 &v4 &Avg. &v1 &v2 &v3 &v4 &Avg. \\
\hline
GLAR &\bf{93.68} &\bf{94.74} &\bf{93.32} &\bf{92.44} &\bf{93.55} &\bf{91.34} &\bf{96.59} &\bf{95.95} &\bf{96.37} &\bf{95.06} &\bf{91.5} &\bf{97.26} &\bf{95.61} &\bf{93.16} &\bf{94.38}\\
\hline
GLAR w/o D&91.37&90.71&89.14&88.57&89.95 &87.56 &95.39 &94.91 &94.99 &93.21 &88.50 &96.33 &93.65 &91.81 &92.57\\
$\Delta$ &-2.31 &-4.03	&-4.18 &-3.87
&-3.60 &-3.78 &-1.20 &-1.04 &-1.38
&-1.85 &-3.00 &-0.93 &-1.96 &-1.35 &-1.81\\
\hline
GLAR w/o R&92.63&93.84&91.55&91.21 &92.31&86.82 &94.56 &92.35 &93.65 &91.85 &87 &95.48 &92.83 &89.19 &91.12\\
$\Delta$ &-1.05 &-0.90 &-1.77 &-1.23
&-1.24 &-4.52 &-2.03 &-3.60 &-2.72
&-3.21 &-4.5 &-1.78 &-2.78 &-3.97 &-3.26\\
\hline
GLAR w/o L&90.59&93.52&91.44 &90.71&91.57&87.89	&95.18 &93.46 &94.52 &92.76 &87.50 &94.85 &92.58 &90.28 &91.30\\
$\Delta$ &-3.09 &-1.22&-1.88 &-1.73
&-1.98 &-3.45 &-1.41 &-2.49 &-1.85 &-2.30 &-4.00 &-2.41 &-3.03 &-2.88 &-3.08\\
\hline
GLAR w/o G&90.36&92.77&92.19&89.97&91.32&89.26 &95.61 &94.17 &94.69	&93.43 &88.5 &95.38 &93.09 &91.12 &92.02\\
$\Delta$ &-3.32 &-1.97 &-1.13 &-2.47
&-2.22 &-2.08 &-0.98 &-1.78 &-1.68&-1.63 &-3.00 &-1.88 &-2.52 &-2.04 &-2.36\\
\bottomrule
\end{tabular}
\label{tab_ablation}
\end{table*}

\begin{table*}[t]
\caption{The time cost (in seconds) of reasoning with different number of candidates on the versions v1, v2, v3 and v4 of inductive test datasets WN18RR-ind, FB15k237-ind and NELL995-ind. "\#Neg." represents the number of negative candidates to be ranked. }
\centering
\setlength{\aboverulesep}{0pt}
\setlength{\belowrulesep}{0pt}
\begin{tabular}{l|l|cc|c||cc|c||cc|c}
\toprule
&\multirow{3}{*}{\#Neg.} & \multicolumn{3}{c||}{WN18RR-ind}  & \multicolumn{3}{c||}{FB15k237-ind} & \multicolumn{3}{c}{NELL995-ind} \\

&& GraIL & GLAR &Speedup& GraIL & GLAR &Seedup &GraIL & GLAR &Speedup \\
 \hline
\multirow{5}{*}{v1}
&20 &33.86 &6.68 &5.1x &93.11 &14.46 &6.4x &28.43 &12.87 &2.2x\\
&50 &49.37 &6.87 &7.2x&173.06 &14.74 &11.7x &46.18 &12.99 &3.6x\\
&80 &56.19 &7.04 &8.0x&253.99 &14.86 &17.1x &63.16 &13.18 &4.8x\\
&120 &73.48 &7.11 &10.3x &328.96 &14.88 &22.1x &76.88 &13.52 &5.7x\\
&150 &85.63 &7.18 &11.9x &401.71 &14.92 &26.9x &99.01  &13.66 &7.2x\\
\hline
\hline
\multirow{5}{*}{v2}
&20 &57.34 &11.12 &5.2x &284.75 &52.73 &5.4x &252.75 &65.49 &3.9x\\
&50 &92.59 &11.17 &8.3x &536.91 &52.95 &10.1x &425.94 &65.82 &6.5x\\
&80 &121.85 &11.39 &10.7x &771.07 &53.06 &14.5x &593.74 &65.97 &9.0x\\
&120 &153.11 &11.53 &13.3x &1055.84 &53.12&19.9x &813.26 &66.24 &12.3x\\
&150 &181.37 &11.75 &15.4x &1382.61 &53.54 &25.8x&961.56 &66.68 &14.4x\\
 \hline
 \hline
\multirow{5}{*}{v3}
&20 &91.32 &34.42 &2.6x &801.67 &158.97 &5.0x &463.54 &159.81 &2.9x\\
&50 &141.18 &34.65 &4.1x &1593.82 &159.54 &10.0x &768.39 &160.24 &4.8x\\
&80 &183.71 &34.87 &5.3x &2292.52 &159.84 &14.3x &996.27 &160.39 &6.2x\\
&120 &239.53 &35.18 &6.8x &3194.07 &160.71 &19.9x &1265.91 &160.52 &7.9x\\
&150 &285.39 &35.39 &8.1x &3821.81 &161.25 &23.7x &1961.56 &160.76 &12.2x\\
\hline
\hline
\multirow{5}{*}{v4}
&20 &205.94 &50.26 &4.1x &1712.35 &384.21 &4.5x &391.48 &144.75 &2.7x\\
&50 &316.78 &50.48 &6.3x &3381.15 &385.36 &8.8x &717.72 &144.83 &5.0x\\
&80 &437.11 &50.64 &8.6x &4839.07 &387.73 &12.5x &982.92 &144.96 &6.8x\\
&120 &588.72 &50.87 &11.6x &6644.15 &388.55 &17.1x &1341.65 &145.23 &9.2x\\
&150 &692.35 &51.07 &13.6x &8033.43 &388.92 &20.7x &1587.43 &145.52 &10.9x\\
\bottomrule
\end{tabular}
\label{tab_reason_time2}
\end{table*}

\subsubsection{Performance on Triple Classification Metric}
Since most previous methods (e.g., GraIL \cite{GraIL}, TACT \cite{TACT}, CoMPILE \cite{CoMPILE}, ConGLR \cite{ConGLR} and etc.) convert inductive KGC task into graph classification task by extracting enclosing subgraph around each triple, the triple classification metric (e.g., AUC-PR) is widely applied to measure the classification performance in these methods. In this paper, we also report the AUC-PR metric on inductive datasets for comparison, as shown in Table~\ref{tab_res_auc}. The AUC-PR results are obtained by taking the mean scores of five runs with different random seeds. 

From Table~\ref{tab_res_auc} we can observe that previous SOTA methods have achieved impressive performance in terms of AUC-PR  (e.g., over 90\% or even close to 100\% AUC-PR scores on majority of these datasets), resulting in less space for any further improvement.  This is because the task of triple classification only requires to classify one negative triple for each positive triple,  which is less challenging than the link prediction task. In the triple classification task, our GLAR model also achieves the best or second best AUC-PR performance on nine out of twelve versions of three datasets and obtains the best average AUC-PR performance on these three datasets. These experimental results again confirm the superiority of our GLAR model.

\subsection{Ablation Study}
In this section, ablation studies are conducted to investigate the contributions of each module of the proposed GLAR model. Specifically, we develop several variant models to validate the impact of different features used in GLAR model: (1) ``GLAR w/o D" denotes the model derived from GLAR by removing the distance feature. (2) "GLAR w/o R" is constructed by removing the relational feature. (3) "GLAR w/o L" removes local structure feature. (4) "GLAR w/o G" discards global structure feature. Table~\ref{tab_ablation} illustrates the experimental results over hits@10 metrics on the three inductive KGC datasets WN18RR-ind, FB15k237-ind and NELL995-ind. We can observe that all variants of the proposed GLAR model achieve worse performance than the original GLAR model, indicating all the components of our GLAR model have a positive contribution to the performance.

Specifically, when removing the distance feature, the hits@10 performances of ``GLAR w/o D" drop a lot on all the three datasets (e.g., averagely drop  3.6\% on WN18RR-ind, 1.85\% on FB15k237-ind and 1.81\% on NELL995-ind). In our GLAR model, the distance feature is designed to measure the closeness between the candidate nodes and center node, which can provide helpful information to predict the relations between entities. From the results of "GLAR w/o R" which removes relational features, we can see that the average hits@10 performances reduce by 1.24\%, 3.21\% and 3.26\% on three datasets, respectively. The reason is that the relational features can convey some important information about the type of entity. For example, the head entity of the relation "{\em PlaceOfBirth}" is always a person. Furthermore, removing local or global structure feature also results in performance reduction on all the datasets, demonstrating that our GLAR model can learn better entity representations by taking full advantage of these structure features.

\begin{figure*}[t]
\centering
\includegraphics[width=0.85\textwidth]{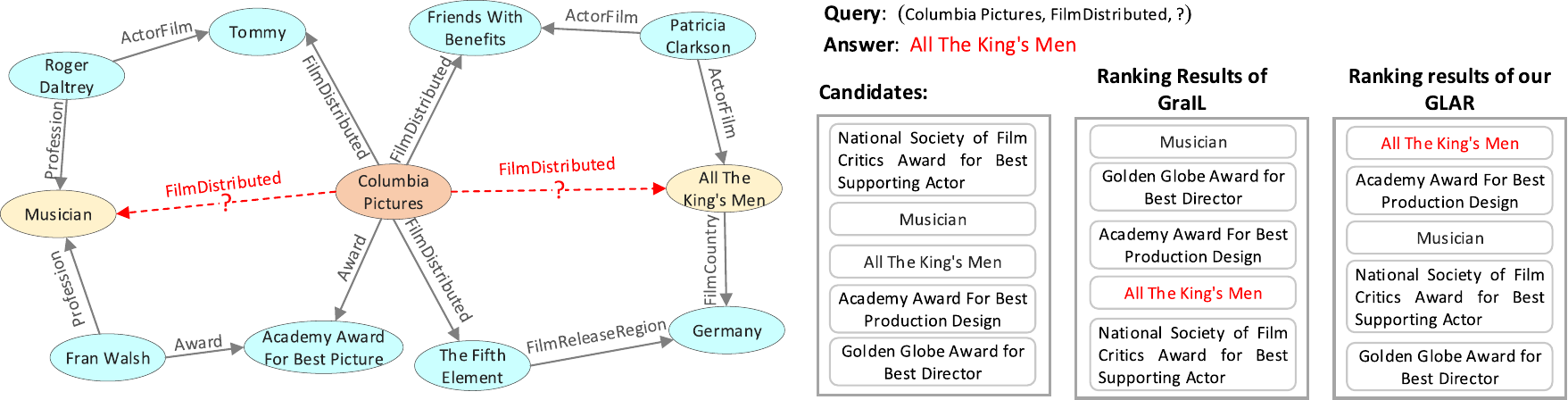}
\caption{The case study regarding the example of link prediction on FB15k237-ind dataset.}
\label{fig_case}
\end{figure*}

\begin{figure*}
    \begin{center}
    \begin{minipage}{0.32\textwidth}
     \includegraphics[width=0.9\columnwidth]{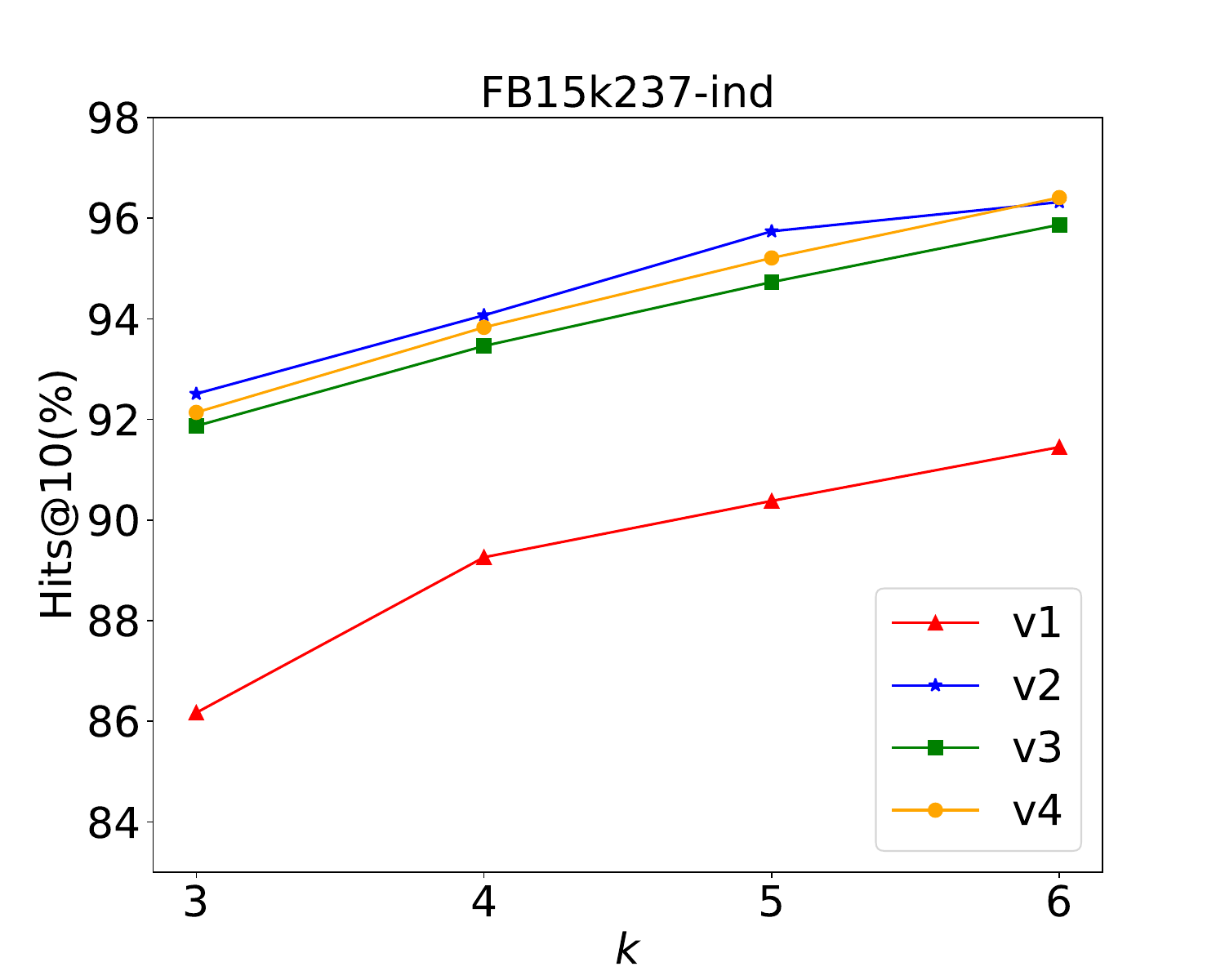}
     \caption{Hits@10 results with different $k$.}
     \label{fig_subgraph_size}
    \end{minipage}
    \begin{minipage}{0.32\textwidth}
     \includegraphics[width=0.9\columnwidth]{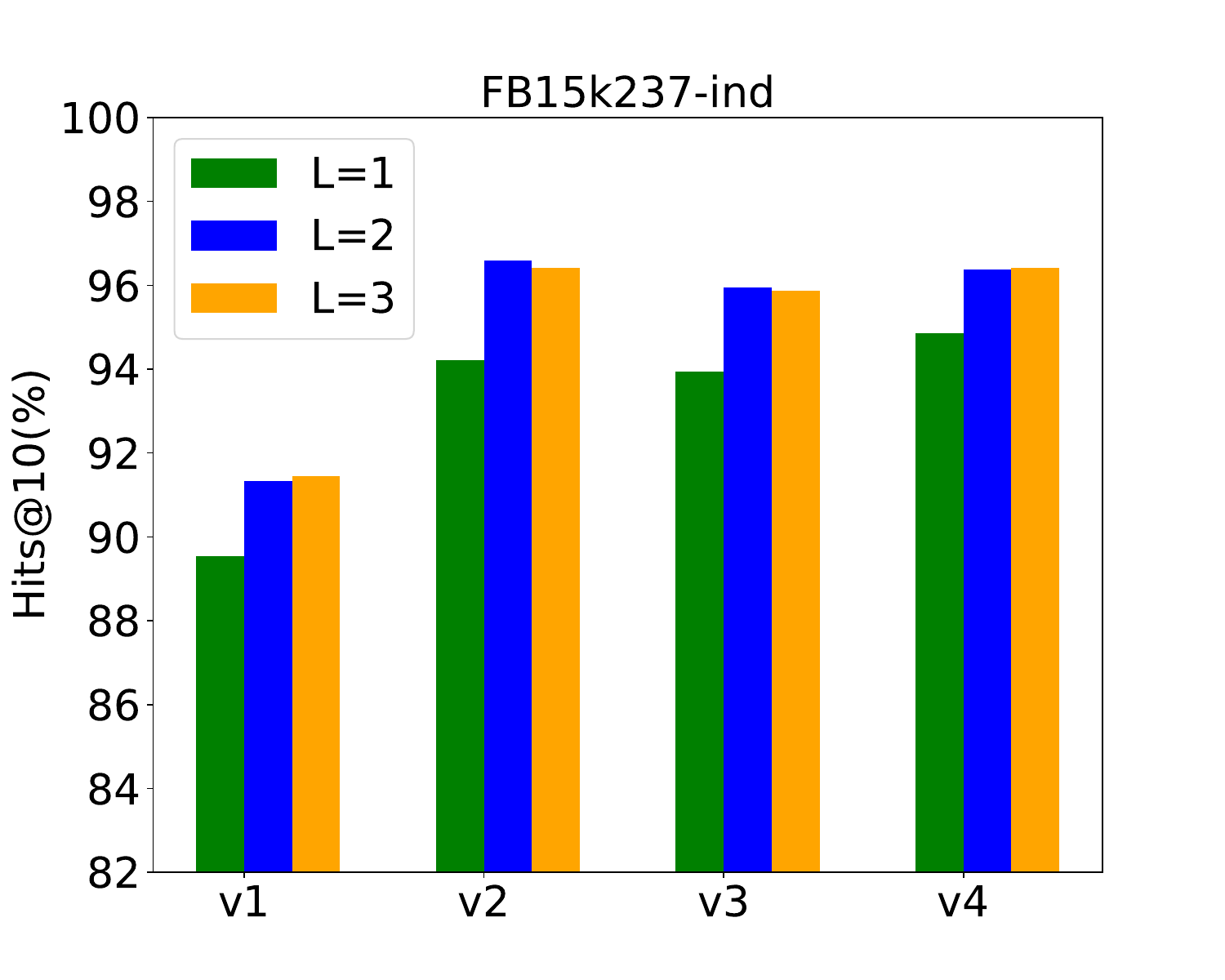}
     \caption{Hits@10 results with different $L$.}
    \label{fig_layer_num}
    \end{minipage}
    \begin{minipage}{0.32\textwidth}
     \includegraphics[width=0.9\columnwidth]{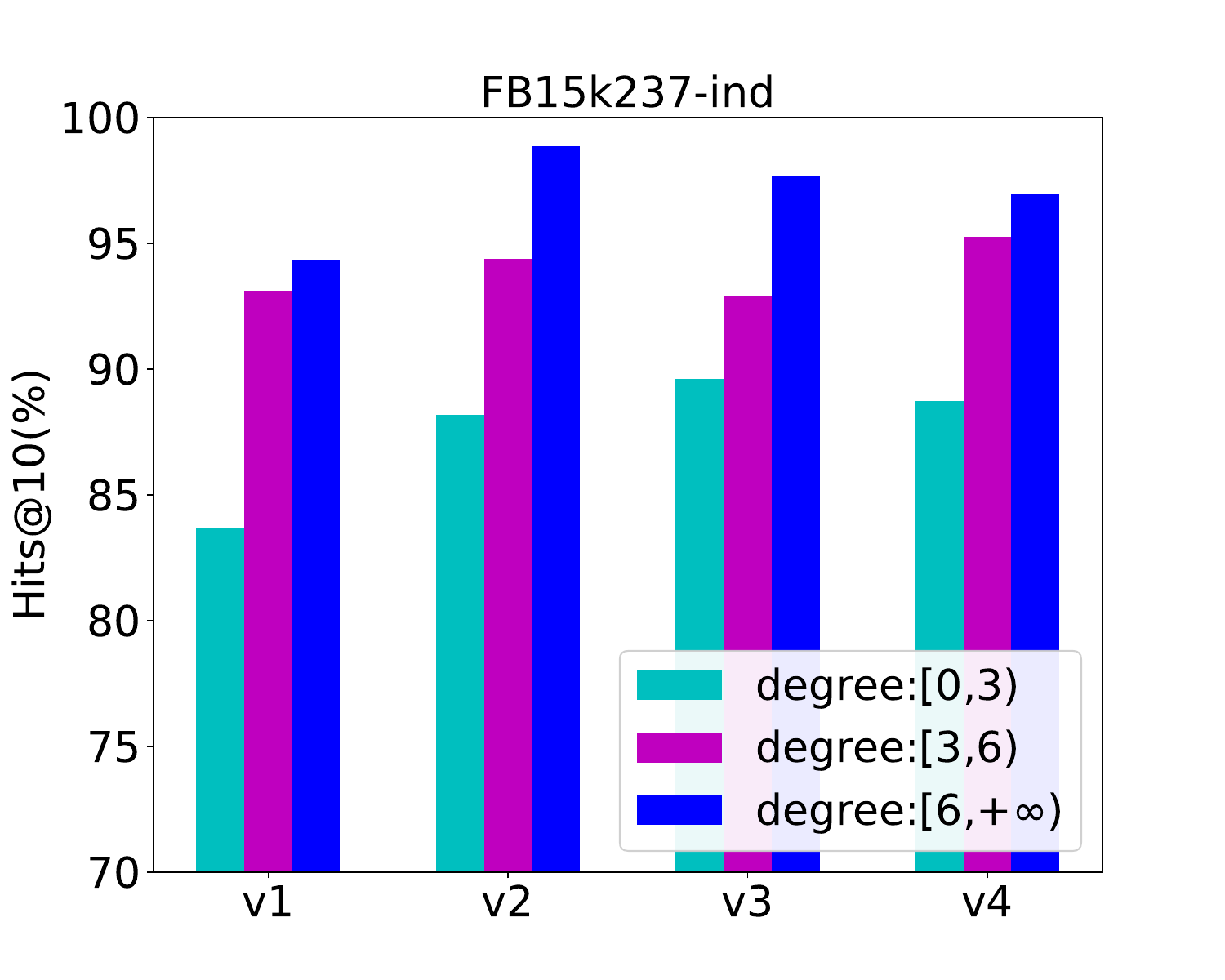}
     \caption{Hits@10 results under different degrees.}
    \label{fig_degree}
    \end{minipage}
    \end{center}
\end{figure*}

\subsection{Efficiency Analysis}
Compared to previous enclosing subgraph based models, we introduce a new opening subgraph based paradigm for inductive KGC, which can perform inductive reasoning with lower cost. To investigate the reasoning efficiency, we compare the reasoning time between our proposed GLAR model and previous enclosing subgraph based model (e.g., GraIL \cite{GraIL}). It's worth noting that there are also other enclosing subgraph based models, such as TACT \cite{TACT},  Meta-iKG \cite{MetaiKG}, CoMPILE \cite{CoMPILE} and ConGLR \cite{ConGLR}. These methods are based on GraIL \cite{GraIL} and introduce some additional module to improve the performance. Therefore, the time complexity of these model are higher or close to GraIL. Hence, the GraIL model is chosen as the representative model for comparison.

Table~\ref{tab_reason_time2} shows the time cost to rank different number of candidates on the versions v1, v2, v3 and v4 of inductive test datasets of WN18RR-ind, FB15k237-ind and NELL995-ind. From Table~\ref{tab_reason_time2} we can observe two interesting findings: (1) our GLAR model can perform reasoning with less time than GraIL in condition of the same number of negative samples. For example, when ranking 50 negative samples, our GLAR model performs about 7.2x, 11.7x and 3.6x faster than GraIL on the v1 version of WN18RR-ind, FB15k237-ind and NELL995-ind, respectively.  (2) The time consumption of the GraIL model increases rapidly as the number of negative samples increases. For example, the GraIL spends 93.11 seconds for reasoning on the v1 version of FB15k237-ind when ranking 20 negative samples, while spends much more time (e.g., 401.71 seconds) when ranking 150 negative samples. In contrast, our GLAR model can consistently maintain low time consumption for different numbers of negative samples. These experimental results indicate that our GLAR model is more efficient than previous methods that use enclosing subgraphs for reasoning.

\subsection{Case Study}
Figure~\ref{fig_case} illustrates an example of link prediction on FB15k237-ind dataset, where the query is ({\em Columbia Pictures}, {\em FilmDistributed},?) and the answer entity is ``{\em All The King's Men}". We let the model rank the answer entity among five candidates and compare the ranking results of GraIL model and our GLAR model. We can see that the GraIL model fails to predict the correct answer entity and ranks the negative candidate entity ``{\em Musician}" at the first position. This is mainly because the GraIL method will extract similar enclosing subgraphs for the entities ``{\em All The King's Men}" and ``{\em Musician}", where only the distances from each node to candidate links are used as node features. Instead, our GLAR model can effectively learn rich node features via entity-independent anchors. This enables our GLAR model to correctly rank the answer entity ``{\em All The King's Men}" at the first position. The case study again verifies the advantages of the proposed GLAR model.

\subsection{Parameter Sensitivity Analysis}
In this subsection, we investigate the impact of hyperparameters in GLAR, including (1) the size of the opening subgraph (e.g., $k$); (2) the number of global-local graph reasoning layers (e.g., $L$).

\subsubsection{The Impact of the Subgraph Size $k$}
The opening subgraph provides useful information for inductive reasoning, therefore the size of the opening subgraph is an important hyperparameter in our GLAR model. We analyze the impact of $k$ on FB15k237-ind dataset in Figure~\ref{fig_subgraph_size}. We can observe that the Hits@10 score gradually increases as the size of the subgraph (e.g., $k$) increases from 3 to 6. This is because larger opening subgraphs can provide local structure information for more entities, thus enabling the model to better rank candidate entities. However, increasing the size of the subgraphs will also lead to increased time and memory consumption. In this paper, we select $k=6$ for the trade-off between performance and overhead.

\subsubsection{The Impact of the Layer Number $L$}
Figure~\ref{fig_layer_num} shows the Hits@10 performance on FB15k237-ind under with difference number of global-local graph reasoning layers (e.g., $L$).  It can be seen that as $L$ increases from 1 to 2, the results on all the four versions gain obvious improvements. This is because using multiple layers can enable our model to capture rich multi-hop structure information. However, the performance cannot be further improved by increasing $L$ to 3, and even slightly declines on the v2 and v3 versions of FB15k237-ind. One potential reason is that stacking more layers will introduce more noise information. Therefore, we set $L=2$ in our experiments.

\subsection{Comparison Across Different Entity Degrees}
In the proposed GLAR model, both the local and global structure features depend on the degree of entities. Therefore, we compare the performance on FB15k237-ind across different degrees of entities to analyze the impact of entity degree. As shown in Figure~\ref{fig_degree}, we divide each test dataset into three groups according to the degree of answer entities. We can find that the proposed GLAR model can achieve good performance on samples with higher entity degree (e.g., degree [3,6) and [6,$+\infty$)), while it performs worse on samples with degree less than 3. The reason is that the GLAR model cannot capture sufficient structure information for entities of small degree, resulting in a performance drop. This indicates that although the proposed GLAR model has achieved good results, it still suffers from a limitation in handling sparse KGs.

\section{Conclusion and Future Work}
\label{sec_conclusion}
In this paper, a novel global-local anchor representation (GLAR) learning model is proposed for the inductive KGC task. Different from previous enclosing subgraph based models, we introduce a new paradigm based on opening subgraph, which is able to rank all candidates for a given query in one subgraph. We design some local and global anchors to learn rich entity-independent structure features for the nodes in KGs. And we use a global-local graph reasoning model to integrate both local and global information to perform inductive reasoning on KGs. We conduct extensive experiments on three widely used inductive KGC datasets and show the superiority of the proposed model in both effectiveness and efficiency.

For future work, we intend to improve our model in three directions. First, we will develop effective subgraph sampling methods to reduce the usage of GPU memory so that our model can be applied to large KGs.  Second, the description information is important and useful for inductive KGC task, thus we plan to improve our method by incorporating text descriptions of entities and relations. Third, this work mainly focuses on the inductive setting for emerging entities, while new relations will also continue to emerge in practice. Therefore, we will design new models to address the inductive KGC setting where both entities and relations are unseen during training.

\ifCLASSOPTIONcompsoc
  \section*{Acknowledgments}
\else
  \section*{Acknowledgment}
\fi

This work was supported by the National Natural Science Foundation of China under Grant 62377021,
financially supported by self-determined research funds of CCNU from the colleges’ basic research and operation of MOE (No. CCNU22QN015), the Natural Science Foundation of Hubei Province for Distinguished Young Scholars (No. 2023AFA096), and theWuhan Knowledge Innovation Project (No.2022010801010278). This work was also supported by the Natural Sciences and Engineering Research Council (NSERC) of Canada, an NSERC CREATE award in ADERSIM\footnote{http://www.yorku.ca/adersim } , the York Research Chairs (YRC) program and an ORF-RE (Ontario Research Fund-Research Excellence) award in BRAIN Alliance\footnote{http://brainalliance.ca}. We would also like to thank all the reviewers and associate editor for their excellent comments.

\ifCLASSOPTIONcaptionsoff
  \newpage
\fi

\bibliographystyle{IEEEtran}
\bibliography{bare_jrnl_compsoc}

\begin{IEEEbiography}[{\includegraphics[width=1in,height=1.25in,clip,keepaspectratio]{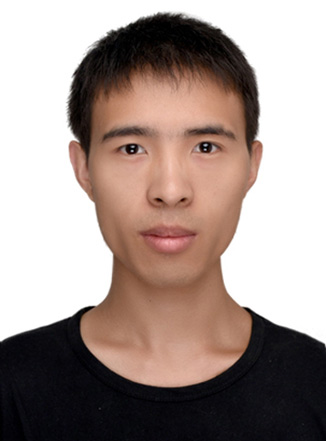}}]{Zhiwen Xie} received the B.S. degree and M.S. degree from Central China Normal University, Wuhan, China, in 2015 and 2018, respectively. And he completed the PhD degree at the School of Computer Science, Wuhan University, Wuhan, China, 2023. He is currently a special associate researcher at Central China Normal University. His research interests include natural language processing, knowledge graph embedding, and information retrieval.
\end{IEEEbiography}
\begin{IEEEbiography}[{\includegraphics[width=1in,height=1.25in,clip,keepaspectratio]{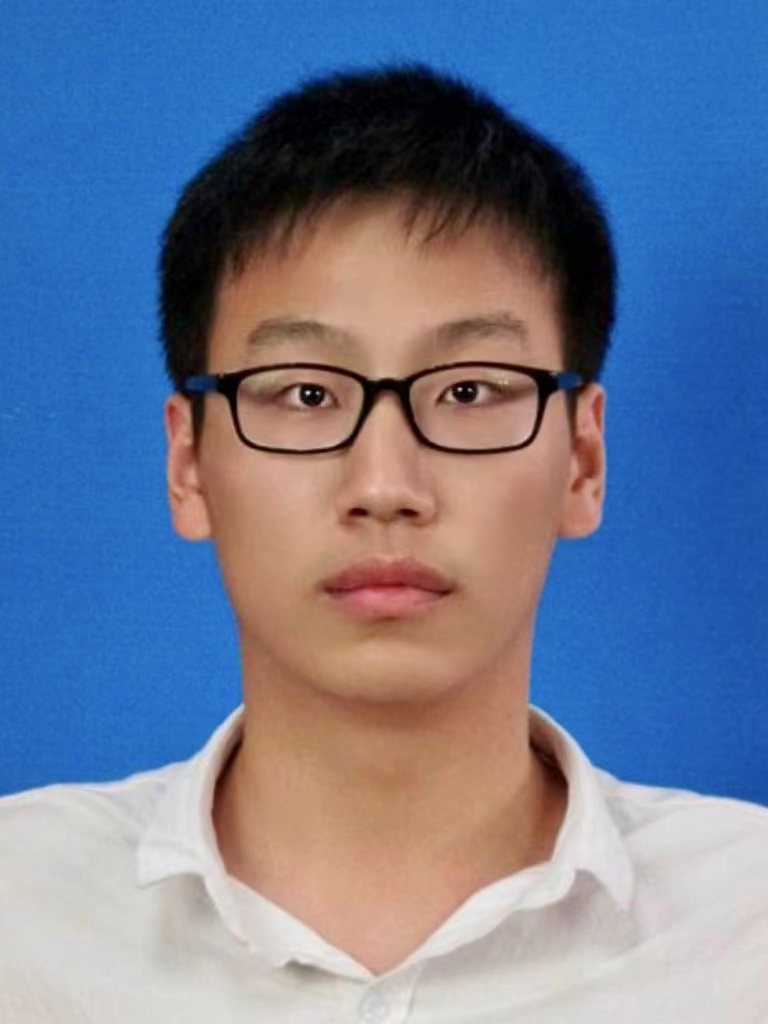}}]{Yi Zhang} received the B.S. degree from Nanchang University, Nanchang, China, in 2016, and the M.S. degree in 2019 from Central China Normal University, Wuhan, China, where he is currently working toward the Ph.D. degree. His research interests include natural language processing and smart education.
\end{IEEEbiography}
\begin{IEEEbiography}[{\includegraphics[width=1in,height=1.25in,clip,keepaspectratio]{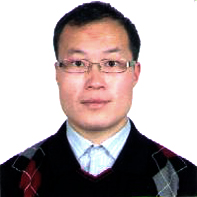}}]{Guangyou Zhou} received the PhD degree from the National Laboratory of Pattern Recognition (NLPR), Institute of Automation, Chinese Academy of Sciences (IACAS), in 2013. Currently, he is working as a professor in the School of Computer, Central China Normal University. His research interests include natural language processing and information retrieval. He has won the best paper award in COLING 2014 and NLPCC 2014. Now, he has served on several program committees of the major international conferences in the field of natural language processing and knowledge engineering, and also served as a reviewer for several journals. Since 2011, he has published more than 40 papers in the leading journals and top conferences, such as the IEEE Transactions on Knowledge and Data Engineering, IEEE Transactions on Cybernetics, ACM Transactions on Information Systems, ACM Transactions on the Web, the IEEE/ACM Transactions on Audio, Speech, and Language Processing, ACL, IJCAI, CIKM, COLING, etc.
\end{IEEEbiography}
\begin{IEEEbiography}[{\includegraphics[width=1in,height=1.25in,clip,keepaspectratio]{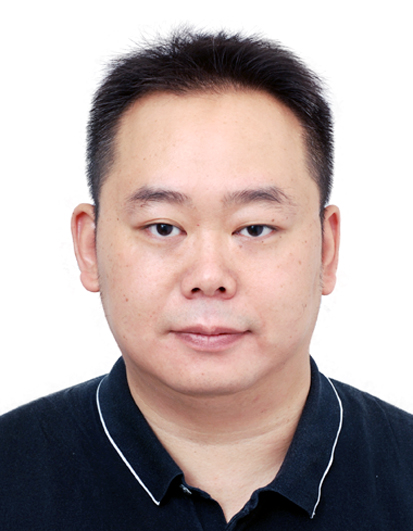}}]{Jin Liu} received his PhD degree from Wuhan University, China, in 2005. He is now a full professor in School of Computer Science, Wuhan University. His main research interests include machine learning and data mining. He has published more than 60 papers in well-known conferences and journals.
\end{IEEEbiography}
\begin{IEEEbiography}[{\includegraphics[width=1in,height=1.25in,clip,keepaspectratio]{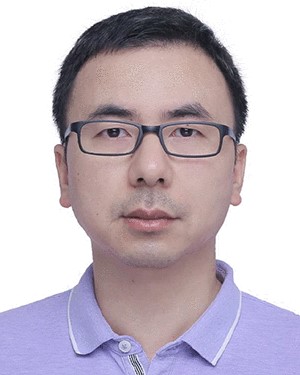}}]{Xinhui Tu} received the B.Sc., M.Sc., and Ph.D., degrees from Central China Normal University, Wuhan, China, in 2001, 2006, and 2012, respectively. He is currently an Associate Professor with the School of Compute, Central China Normal University. His research interests include information retrieval and natural language processing.
\end{IEEEbiography}

\begin{IEEEbiography}[{\includegraphics[width=1in,height=1.25in,clip,keepaspectratio]{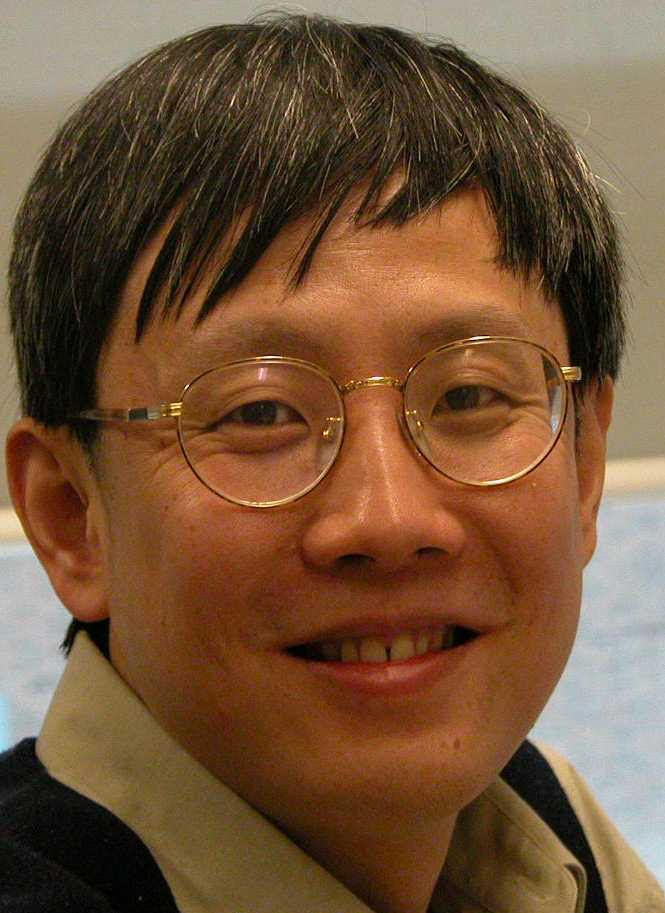}}]{Jimmy Xiangji Huang} received the Ph.D. degree in information science from the City, University of London, United Kingdom, and was then a Postdoctoral Fellow at the School of Computer Science, University of Waterloo, Canada. He is now a Tier I York Research Chair (YRC) professor and the Director of the Information Retrieval \& Knowledge Management Research Lab (IRLab), York University. He joined York University as an assistant professor in 2003. He was early-tenured in 2006, promoted to Full Professor in 2011 and awarded as York Research Chair in 2016 respectively. He received the Dean’s Award for Outstanding Research in 2006, an Early Researcher Award, formerly the Premiers Research Excellence Award in 2007, the Petro Canada Young Innovators Award in 2008, the SHARCNET Research Fellowship Award in 2009, the Best Paper Award at the 32nd European Conference on Information Retrieval (ECIR 2010) and LA\&PS Award for Distinction in Research, Creativity, and Scholarship (established researcher) in 2015. He has been selected as a 2024 York University Research Award Winner. Since 2003, he has published over 320 refereed papers in top-tier journals (such as ACM Transactions on Information Systems, IEEE Transactions on Knowledge and Data Engineering, IEEE/ACM Transactions on Audio, Speech and Language Processing, ACM Transactions on Intelligent Systems and Technology, ACM Computing Surveys, Information Sciences, Journal of American Society for Information Science and Technology, Artificial Intelligence in Medicine, Journal of Computers in Biology and Medicine, Computational Linguistics and BMC Genomics) and major conferences in the fields (such as ACM SIGIR, ACM CIKM, ACM SIGKDD, ACL, EMNLP, COLING, IJCAI \& AAAI). He was the General Conference Chair for the 19th International ACM CIKM Conference and the General Conference Chair for the 43rd International ACM SIGIR Conference. He is a senior member of the IEEE and an ACM Distinguished Scientist.
\end{IEEEbiography}

\end{document}